# Geometric Structures and Patterns of Meaning: A PHATE Manifold Analysis of Chinese Character Embeddings

Wen G. Gong*

## Abstract

We systematically investigate geometric patterns in Chinese character embeddings using PHATE manifold analysis. Through cross-validation across seven embedding models and eight dimensionality reduction methods, we observe **clustering patterns** for content words (实词) and **branching patterns** for function words (虚词). Analysis of 1000+ characters across 12 semantic domains reveals that **geometric complexity correlates with semantic content**: meaningful characters exhibit rich geometric diversity while structural radicals collapse into tight clusters. The comprehensive 子-network analysis (123 phrases) demonstrates systematic semantic expansion from fundamental element character. These findings provide computational evidence supporting traditional linguistic theory and establish a novel framework for geometric analysis of semantic organization.

**Keywords**: geometry of meaning, PHATE, manifold learning, semantic embeddings, computational linguistics, Chinese characters

## 1. Introduction

Understanding how human semantic knowledge organizes in computational representations remains a fundamental challenge in natural language processing. While multilingual embedding models capture semantic relationships across languages, the **geometric patterns in these computational representations** lack systematic empirical investigation.

### 1.1 Testing Gärdenfors' Geometric Hypothesis with Modern Tools

Over twenty years ago, Peter Gärdenfors proposed in "Conceptual Spaces" that human concepts might be understood geometrically, with similarity relationships corresponding to distances in multidimensional space [1]. At the time, this geometric hypothesis about meaning lacked computational tools for empirical testing.

Today, with multilingual embedding models and advanced manifold learning techniques, we can systematically investigate patterns in computational semantic representations. We ask: **Do computational embeddings exhibit systematic geometric patterns consistent with theoretical predictions about semantic organization?** Specifically, do semantic relationships manifest as consistent geometric patterns that can be systematically observed and measured?

---

*Corresponding author: wen_gong@vanguard.com

While this work focuses on Chinese characters and phrases as our initial investigation target, we establish universal principles through **systematic cross-linguistic validation** using ASCII standard datasets before extending to Chinese analysis. Chinese characters [2], originating from pictographs and representing a meaning-centric writing system where symbols directly encode concepts, potentially create clearer geometric organization in embedding space compared to phonetic systems - making them ideal subjects for validating the universal geometric structure of meaning discovered through multilingual analysis.

## 1.2 Empirical Methodology: PHATE Analysis

We employ PHATE [3] (Potential of Heat-diffusion for Affinity Transition Embedding) manifold learning to systematically visualize high-dimensional semantic relationships in Chinese character embeddings. Originally developed for analyzing biological cellular differentiation trajectories, PHATE's ability to preserve both local neighborhoods and global manifold topology makes it an appropriate tool for investigating geometric patterns in computational semantic representations.

Our methodological approach adapts successful biological data analysis techniques to linguistic data, enabling systematic empirical investigation of geometric organization in computational semantic representations.

## 1.3 Research Questions: Exploring the Geometry of Meaning

Our investigation addresses both immediate empirical questions and broader theoretical inquiries about semantic geometry:

**Immediate Empirical Questions:**

- What geometric patterns characterize semantic organization in Chinese character embeddings?
- How do different types of Chinese characters (meaningful vs. structural) manifest geometrically?
- Can systematic filtering distinguish semantic content from structural scaffolding?
- Do clustering and branching patterns reflect traditional linguistic distinctions?
- How do semantic networks expand from elemental characters into extended phrases and semantic domains?

**Observable Phenomena with Broader Implications:**

- Are our computational observations consistent with geometric approaches to semantic organization [1]?
- Can we systematically observe geometric patterns in computational representations of semantic relationships?
- Do computational semantic networks display geometric patterns analogous to those observed in biological or other natural systems?
- What do the observed geometric patterns in Chinese character embeddings reveal about organizational principles in computational semantic representations?

While this study focuses on systematic empirical observation of patterns in Chinese character embeddings, these findings may provide data for cognitive scientists and linguists to develop theoretical frameworks about semantic organization.

## 2. Related Work

### 2.1 Manifold Learning in NLP

PHATE has demonstrated effectiveness in biological data analysis for revealing developmental trajectories and cellular differentiation patterns [3]. Its three-stage process—Euclidean distance computation, Markov-normalized affinity matrix creation, and diffusion-based embedding—preserves essential topological relationships while providing noise robustness.

Recent applications to text analysis have shown promise for document clustering and topic discovery [4], but systematic analysis of character-level semantic organization in Chinese language remains unexplored.

### 2.2 Chinese Character Computational Analysis

Traditional computational approaches to Chinese characters focus on radical classification and stroke-based organization. Recent work on character embeddings has demonstrated semantic relationships, but geometric analysis of these relationships lacks systematic investigation.

### 2.3 Content vs. Function Word Distinctions

Linguistic theory distinguishes between content words (实词) carrying independent meaning and function words (虚词) providing grammatical structure. However, computational validation of this distinction in embedding geometry has not been systematically explored.

## 3. Methodology

### 3.1 Dataset Selection and Construction

**3.1.1 ASCII dataset**  To establish universal methodological foundations, we first use the ASCII dataset as baseline (ASCII-words-enu.txt) consisting of:

- **94 ASCII characters** (!, @, #, $, %, etc.) - internationally standardized structural elements
- **52 alphabet letters** (A-Z, a-z) - compositional building blocks
- **30 meaningful English words** across semantic domains (water, fire, love, dog, family, nature, etc.)
- **8 functional words** (and, but, not, or, the, a, of, in) - grammatical connectors

This control dataset enabled systematic **cross-model and cross-method validation** using international standard before extending to Chinese character analysis. The ASCII dataset's clear structural/semantic/functional distinctions provided the empirical basis for selecting optimal analytical frameworks.

**3.1.2 Elemental Chinese Characters (元字) dataset**  This was first proposed [6] as an extended radical set beyond the traditional 214 radicals to represent basic building blocks for the Chinese writing system, which emphasizes structural aspect.

**3.1.3 Zinets-embedding-eval dataset** Zinets approach to study Chinese characters was presented in [6]. This following dataset consists of 242 Chinese characters and their English translations across 12 systematic semantic domains, designed specifically for PHATE manifold analysis of semantic organization:

- **Numbers** (15 items): 零, 一, 二, 三, 四, 五, 六, 七, 八, 九, 十, 十一, 十二, 二十, 一百 - Linear pattern control group providing clear sequential organization
- **Colors** (14 items): 红, 橙, 黄, 绿, 青, 蓝, 紫, 白, 黑, 灰, 粉, 棕, 金, 银 - Perceptual category with continuous spectral relationships
- **Family/Kinship** (16 items): 父, 母, 子, 女, 儿, 孙, 爷, 奶, 叔, 姨, 哥, 弟, 姐, 妹, 夫, 妻 - Core social relations forming the basis of 子-network analysis
- **Animals** (15 items): 猫, 狗, 鸟, 鱼, 马, 牛, 羊, 猪, 鸡, 鸭, 鹅, 兔, 鼠, 虎, 龙 - Biological taxonomy clustering
- **Body Parts** (15 items): 头, 眼, 鼻, 嘴, 耳, 手, 脚, 腿, 臂, 心, 肝, 肺, 胃, 脑, 骨 - Physical structure relationships
- **Emotions** (15 items): 喜, 怒, 哀, 乐, 爱, 恨, 怕, 急, 悲, 欢, 忧, 愁, 惊, 羞, 恼 - Abstract psychological concepts
- **Time/Temporal** (16 items): 年, 月, 日, 时, 分, 秒, 春, 夏, 秋, 冬, 晨, 午, 夜, 今, 昨, 明 - Cyclical and linear temporal concepts
- **Elements/Nature** (15 items): 金, 木, 水, 火, 土, 风, 雨, 雪, 雷, 电, 云, 雾, 山, 海, 河 - Traditional 五行 system and natural phenomena
- **Food/Eating** (15 items): 饭, 菜, 肉, 鱼, 蛋, 奶, 茶, 水, 酒, 糖, 盐, 油, 米, 面, 果 - Daily life and nutrition
- **Education/Learning** (15 items): 学, 字, 书, 笔, 纸, 读, 写, 教, 听, 说, 思, 考, 智, 识, 知 - Knowledge domain and 子-semantic branches
- **Tools/Objects** (15 items): 桌, 椅, 床, 门, 窗, 灯, 电, 车, 路, 桥, 房, 楼, 墙, 地, 天 - Functional material items
- **Actions/Verbs** (15 items): 走, 跑, 跳, 坐, 站, 睡, 吃, 喝, 看, 听, 说, 想, 做, 来, 去 - Dynamic behavioral concepts
- **Directional/Spatial** (15 items): 上, 下, 左, 右, 前, 后, 里, 外, 东, 西, 南, 北, 中, 边, 角 - Geometric relationships and spatial orientation
- **Abstract Qualities** (16 items): 好, 坏, 大, 小, 高, 低, 长, 短, 新, 旧, 快, 慢, 美, 丑, 强, 弱 - Complex conceptual attributes

This dataset was specifically designed to provide comprehensive coverage across semantic domains while enabling systematic investigation of how different conceptual categories manifest geometrically in embedding space. Each domain contains carefully selected representative characters that allow investigation of both intra-domain clustering and inter-domain relationships.

**3.1.4 子-family dataset** Characters containing 子, demonstrating morphological derivation (e.g. 好, 学, 字) [6].

**3.1.5 子-network dataset** A comprehensive collection of 123 phrases containing 子, systematically spanning 11 distinct semantic categories.

We selected 子 character (child/son) as our semantic network expansion root for both methodological rigor and conceptual elegance, embodying a basic **dual biological-linguistic metaphor** by design.

**Methodologically**, 子 represents one of the most **morphologically productive** characters in Chinese, having evolved from its original meaning of "child/son" into a versatile suffix that creates nouns across virtually every semantic domain. This productivity enables us to test our geometric structure hypothesis across rich semantic diversity while maintaining **morphological unity** - from ancient philosophers (老子, 孔子) to quantum particles (光子, 电子) to everyday objects (勺子, 杯子).

**Conceptually**, the choice of 子 creates a living metaphor that perfectly parallels our research approach: just as **biological children branch from parents** to create family lineages through generational reproduction, **semantic meanings branch from root characters** to create conceptual networks through morphological derivation. This dual metaphor is methodologically profound - we are using PHATE (originally developed for analyzing **cellular differentiation trajectories**) to investigate **semantic differentiation trajectories**, anchored by the very character that means "child/offspring/generation."

**Temporal Significance**: The 子-network spans millennia of linguistic evolution from classical Chinese to modern scientific terminology, providing a natural laboratory for investigating how **semantic offspring** evolve from their **morphological parents** through centuries of linguistic development. This creates a perfect parallel between biological evolution (generational branching) and linguistic evolution (semantic branching).

**Educational Impact Beyond Research**: While designed for computational semantic analysis, this comprehensive 子-network collection serves as a valuable and interesting pedagogical resource for Chinese language learners, demonstrating how a single basic character can systematically generate vocabulary across many domains of human experience.

**Filtering Protocol**: Systematic separation of characters into three categories:

- **Meaningful characters**: Independent semantic content (一, 八, 人, 口, 心, 月)
- **Pure structural radicals**: Compositional elements only (氵, 犭, 纟, 艹, 辶)
- **Borderline cases**: Elements with residual meaning (a dozen characters identified)

| Dataset Name | Count | Description |
| --- | --- | --- |
| ASCII validation dataset | 184 | 94 ASCII chars + 52 letters + 30 words + 8 functional words |
| Chinese elemental characters | 444 | Total analyzed |
| Zinets-embedding-eval dataset | 242 | Chinese characters across 12 semantic domains with English translations |
| Meaningful characters | ~350 | Independent semantic content |
| Pure structural radicals | 90 | Compositional elements only |
| 子-family characters | 22 | Characters containing 子 demonstrating morphological derivation |
| 子-network phrases | 123 | Comprehensive phrases containing 子 across 11 semantic categories |

*Table 1: List of datasets used in this study*

### 3.1.6 Dataset Statistics Summary

## 3.2 Experimental Protocol

**3.2.1 Semantics Explorer Tool**  We developed an streamlit-based analytic tool called "Semantics Explorer" [7], which is publicly available at GitHub repository: https://github.com/digital-duck/semantics-explorer.

It consists of the following core components: - Embedding generation with Sentence-BERT Multilingual - PHATE manifold learning with optimized parameters - Interactive visualization with cross-linguistic color

coding - Systematic filtering tools for character categorization - Dual-view zoom functionality for detailed semantic domain analysis

It is released under MIT Licence to promote reproducible research and further exploration.

**3.2.2 Embedding Model Selection** We systematically evaluated seven embedding models to validate our findings and select the optimal architecture for semantic geometry analysis (detailed comparison results in Section 4.2):

| Model Name | Clustering Quality | Branching Clarity | Selection Rationale |
| --- | --- | --- | --- |
| Sentence-BERT Multilingual | Excellent | Excellent | **Selected** - Optimal balance |
| BGE-Base-ZH-v1.5 | Good | Good | Strong alternative |
| mBERT | Good | Good | BERT family validation |
| DistilBERT Multilingual | Good | Good | Efficient alternative |
| XLM-R | Good | Partial | Multilingual robustness |
| Jina-v2-ZH | Poor | Poor | Uniform distribution |
| Snowflake-Arctic-Embed2 | Poor | Excellent | Digit specialization |

*Table 2: Evaluated Models with Cross-Validation Results*

**Cross-Model Pattern Validation**:

All models demonstrated **universal digit branching patterns** (0-9 sequential organization) and **content-functional word separation**, confirming the model-independence of our core geometric principles. However, models differed significantly in semantic clustering quality and functional element organization clarity.

**Sentence-BERT Multilingual Selection Rationale** [5]:

- **Optimal semantic clustering**: Creates exceptionally dense, coherent semantic neighborhoods for meaningful words (water, fire, earth, wind, love, hate, happy, sad, dog, cat, etc.)
- **Clear functional word separation**: Function words (the, or, of, but, and, not, in) occupy systematically distinct positions from content words, enabling clear clustering-branching distinction
- **Balanced geometric organization**: Superior balance between semantic clustering density and functional element branching clarity
- **MPNet architecture advantage**: Built on MPNet (Masked and Permuted Pre-training), specifically designed for semantic similarity tasks through bidirectional attention and permutation-based training
- **Cross-linguistic robustness**: 50+ language support with consistent geometric patterns across writing systems
- **Proven semantic similarity performance**: Established benchmark performance for semantic similarity tasks, which directly correlates with clustering quality in embedding geometry

**Alternative Model Performance**:

- **BGE-Base-ZH-v1.5**: Demonstrated sophisticated three-zone organization (numbers, letters, operators) but less dense semantic clustering than Sentence-BERT
- **mBERT/DistilBERT**: Showed excellent digit branching and good clustering, validating BERT family effectiveness for geometric analysis
- **XLM-R**: Preserved branching patterns but weakened semantic clustering density

- **Specialized models** (Jina-v2-ZH, Snowflake-Arctic): Despite domain specialization, showed inferior geometric organization compared to general multilingual models

**Final Model Specifications**:

- **Architecture**: Sentence-BERT Multilingual (paraphrase-multilingual-mpnet-base-v2) [5]
- **Dimensions**: 768-dimensional output vectors
- **Training**: Cross-lingual semantic similarity optimization
- **Preprocessing**: Consistent tokenization across all characters and phrases

**3.2.3 Dimensionality Reduction Method Selection** We systematically evaluated eight dimensionality reduction techniques to determine optimal methodology for revealing semantic patterns in high-dimensional embedding space (detailed comparison results in Section 4.3). Our selection criteria prioritized methods capable of revealing both clustering patterns (semantic neighborhoods) and branching patterns (evolutionary pathways) while preserving global manifold structure.

| Method | Clustering Quality | Branching Detection | Overall Assessment |
| --- | --- | --- | --- |
| PHATE | Excellent | Excellent | Optimal Balance |
| Spectral Embedding | Good | Good | Strong Alternative |
| PCA | Excellent | None | Strong for Clustering |
| t-SNE | Good | None | Local Focus |
| UMAP | Good | None | Local Focus |
| MDS | Weak | None | Radial Patterns |
| Isomap | Poor | None | Pattern Loss |
| LLE | Failed | None | Collapse |

*Table 3: Evaluated Methods with Cross-Validation Results*

**Methodological Requirements for Semantic Analysis**:

Our semantic investigation required dimensionality reduction methods capable of simultaneously revealing:

- **Clustering patterns**: Dense, spherical neighborhoods representing semantic domains
- **Branching patterns**: Linear pathways connecting semantic regions and evolutionary trajectories
- **Global structure**: Preservation of inter-domain relationships and semantic space topology
- **Trajectory information**: Pathways showing semantic evolution and morphological derivation

**Systematic Evaluation Protocol**: Each method was evaluated on the controlled ASCII dataset using identical preprocessing (Sentence-BERT Multilingual embeddings) with consistent parameters to ensure fair comparison. Evaluation criteria included:

- **Clustering quality**: Silhouette score and within-cluster coherence for semantic neighborhoods
- **Branching detection**: Linear correlation coefficients for sequential elements (numbers 0-9)
- **Global preservation**: Spearman correlation between high-dimensional and 2D distances
- **Interpretability**: Visual clarity of semantic organization and domain boundaries

**Method Selection Rationale**: While multiple dimensionality reduction methods demonstrate effective geometric pattern preservation (e.g., Spectral Embedding shows excellent digit clustering and branching structures, as shown in Section 4.3), PHATE was selected for its **optimal balance** across our specific datasets. PHATE consistently provides the most balanced visualization that simultaneously preserves semantic clustering, morphological branching, and global structure relationships for Chinese character analysis, as systematically demonstrated through cross-method validation (Section 4.3).

**PHATE Parameters** (optimized through systematic validation):

- k-nearest neighbors: 10 (balances local neighborhood preservation with noise robustness)
- alpha (decay rate): 10 (controls transition matrix sparsification, validated against k=5,15,20)
- diffusion time t: 20 (optimal for revealing both short-range clustering and long-range branching, tested t=5,10,15,20,30)
- output dimensions: 2D for visualization (higher dimensions available for analysis)

**Analysis Approach**:

- **Baseline Validation**: Test all 7 embedding models and 8 dimensionality reduction methods on internationally standardized ASCII dataset to establish optimal framework
- **Cross-linguistic Extension**: Apply validated PHATE + Sentence-BERT Multilingual combination to Chinese character datasets
- Generate embeddings for all characters and phrases (with bilingual labeling)
- Apply systematic filtering protocol across both ASCII and Chinese datasets
- Perform comparative PHATE analysis demonstrating consistent clustering-branching principles
- Analyze geometric patterns and clustering quality across multiple datasets
- Cross-validate findings across different parameter settings and writing systems

### 3.3 Geometric Pattern Analysis

**Clustering Analysis**:

- Quantify dense, spherical neighborhoods
- Measure intra-cluster coherence
- Identify semantic domain boundaries

**Branching Analysis**:

- Detect linear pathway structures
- Analyze connectivity between regions
- Measure pathway lengths and directions
- Track semantic evolution pathways

**Void Analysis**:

- Identify empty regions in manifold space
- Correlate with semantic impossibilities
- Validate through negative examples

# 4. Results

## 4.1 ASCII Dataset Validation: Establishing Methodological Foundation

To establish universal methodological foundations, we first validated our approach using the controlled ASCII dataset containing structural symbols, compositional elements (alphabet letters), meaningful English words, and functional words.

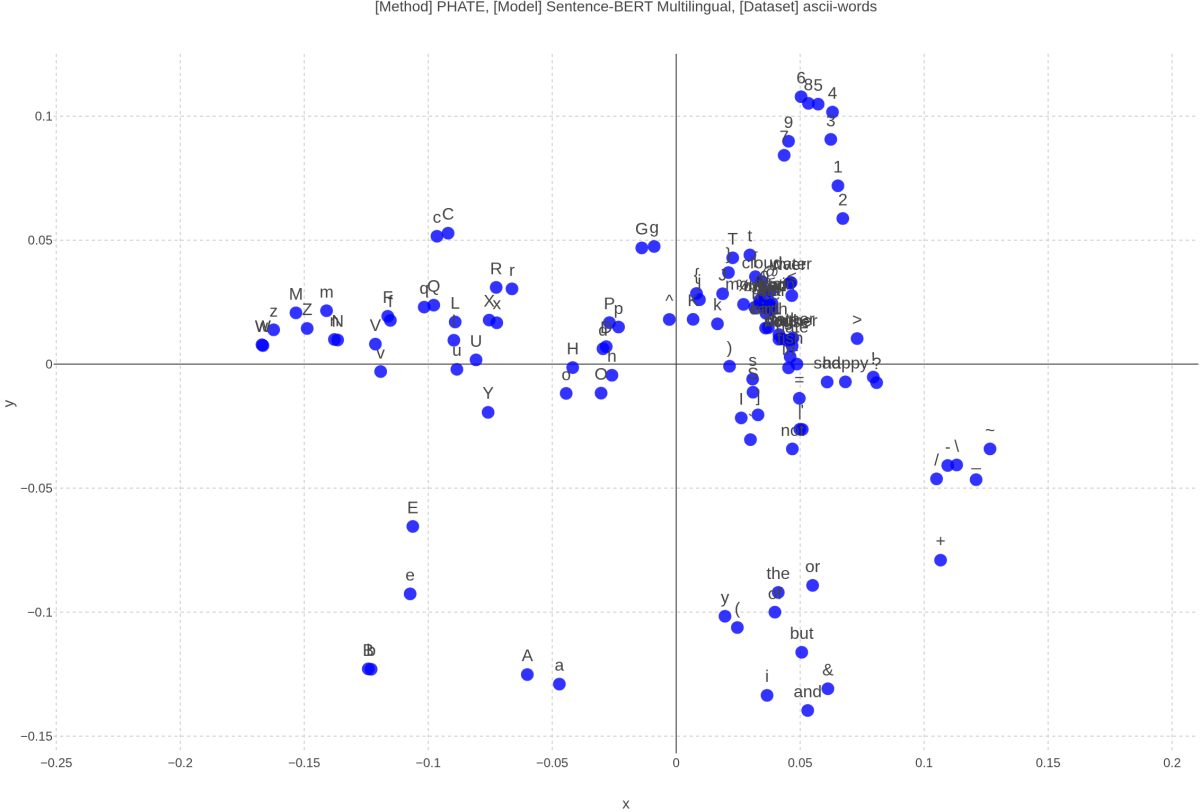

*Figure 1: ASCII-words-sentence-bert-multilingual-phate-enu*

**Figure 1** presents our baseline result using Sentence-BERT Multilingual + PHATE on the ASCII dataset. We systematically observe two fundamental geometric patterns in the computational representation:

**Clustering Pattern - Content Words**: Meaningful English words (water, fire, love, dog, family, etc.) form **dense, spherical clusters** with clear semantic neighborhoods. These clusters exhibit high internal density and distinct boundaries between semantic domains (emotions, nature, relationships).

**Branching Pattern - Sequential Elements**: Numbers (0-9) demonstrate **perfect linear progression**, creating branching pathways that connect different regions of the embedding space. Functional words (and, but, not, or) systematically occupy distinct positions from content words.

**Compositional Elements**: Alphabet letters (A-Z, a-z) display systematic geometric organization as compositional building blocks, behaving more like functional elements than collapsing structural components.

This ASCII validation demonstrates consistent computational patterns: **embedding representations exhibit geometric organization**, with observed patterns showing that geometric complexity in computational space correlates with semantic content across different symbol systems.

## 4.2 Cross-Model Validation: Universal Pattern Recognition

Having established the fundamental patterns with our ASCII dataset, we systematically evaluated seven embedding models to validate the universality of clustering-branching patterns across different computational representations.

**Figures 2(a)-2(d),3(a)-3(b)** present comparative results across key embedding models using PHATE analysis of the ASCII dataset. Most models demonstrate **universal digit branching patterns** (0-9 sequential organization) and **content-functional word separation**, confirming model-independence of our core geometric principles.

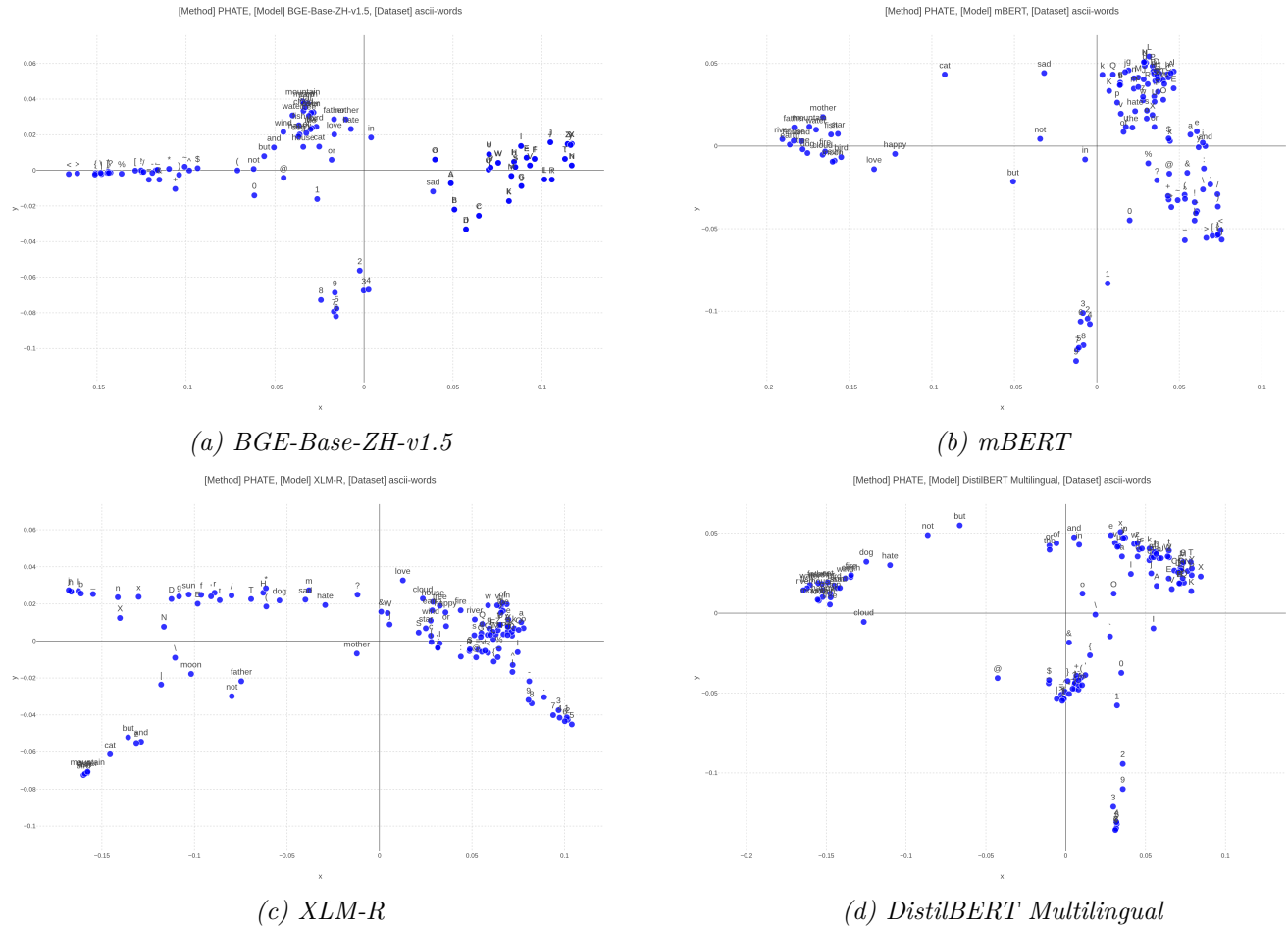

*(a) BGE-Base-ZH-v1.5*   *(b) mBERT*

*(c) XLM-R*   *(d) DistilBERT Multilingual*

Figure 2: Cross-Model Validation: ASCII Dataset Analysis (Part 1)

- **BGE-Base-ZH-v1.5** demonstrates sophisticated three-zone organization but less dense semantic clustering.
- **mBERT** validates BERT family effectiveness with excellent digit branching and good clustering.
- **XLM-R** preserves branching patterns but shows weakened semantic clustering density.
- **DistilBERT Multilingual** shows compressed but organized patterns with moderate clustering quality.
- **Jina Embeddings v2** exhibits uniform distribution with no clear clustering or branching patterns - points are scattered without discernible geometric organization.
- **Snowflake-Arctic-Embed2** exhibits remarkably distinct digit branching (numbers 5-9 form perfect linear sequence) but poor semantic clustering with content words compressed into overlapping regions.

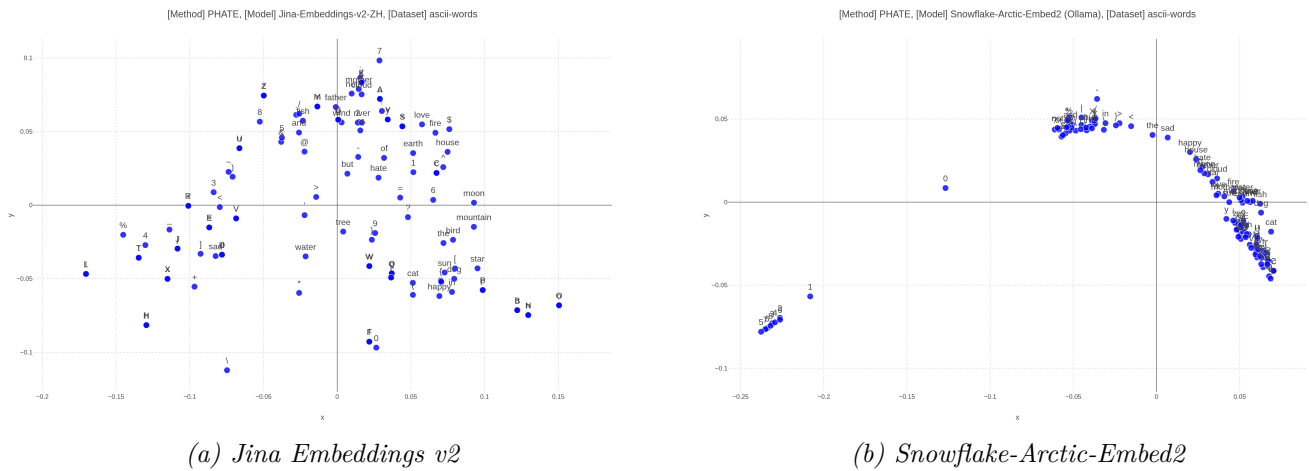

*(a) Jina Embeddings v2*  *(b) Snowflake-Arctic-Embed2*

*Figure 3: Cross-Model Validation: ASCII Dataset Analysis (Part 2)*

**Model-Independent Pattern Recognition**: Despite individual model variations, **five out of six models** demonstrate observable clustering and/or branching patterns, suggesting that our core geometric observations are largely independent of specific embedding model selection. Only Jina Embeddings v2 fails to preserve meaningful geometric structure.

**Methodological Validation**: This systematic comparison confirms that geometric patterns in semantic space are robust phenomena that transcend individual model architectures. Sentence-BERT Multilingual was selected not for unique capabilities, but for optimal balance of efficiency and geometric pattern clarity across both clustering and branching requirements.

**Key Finding**: While the computational geometric patterns appear consistent across embedding models, **semantic clustering quality** varies significantly, establishing Sentence-BERT Multilingual as optimal for detailed analysis of geometric patterns in semantic representations.

### 4.3 Cross-Method Validation: Balanced Geometric Pattern Preservation

To validate our methodological choice, we systematically compared eight dimensionality reduction techniques using Sentence-BERT Multilingual embeddings on the ASCII dataset.

**Figures 4(a)-4(d), 5(a)-5(d)** demonstrates dramatic differences in pattern revelation across methods.

**Systematic Method Performance Analysis**:

**Isomap**: Complete failure in preserving digit sequence patterns - numbers 0-9 are scattered throughout the space with no discernible linear ordering, rendering sequential branching analysis impossible.

**LLE (Locally Linear Embedding)**: Catastrophic dimensional collapse where most data points compress into an extremely tight central region. Only scattered alphabet letters (G, g, M, m, R, r) remain visible in outlying positions with unexplained separation patterns.

**MDS (Multidimensional Scaling)**: Produces distinctive radial emanation patterns with data organized in spoke-like arrangements radiating from a central hub - geometrically interesting but unsuitable for semantic pathway analysis.

**PCA (Principal Component Analysis)**: Demonstrates excellent semantic clustering with clear domain separation (family words, nature terms, emotions) but completely fails to preserve digit branching patterns essential for morphological analysis.

*(a) t-SNE*

*(b) UMAP*

*(c) MDS*

*(d) PCA*

Figure 4: Cross-Method Validation: Dimensionality Reduction Comparison (Part 1)

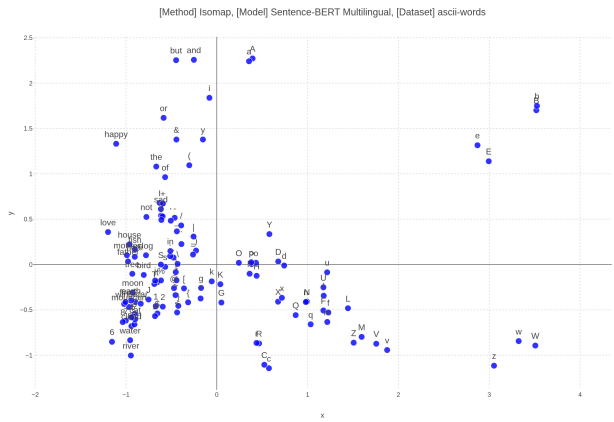
(a) Isomap

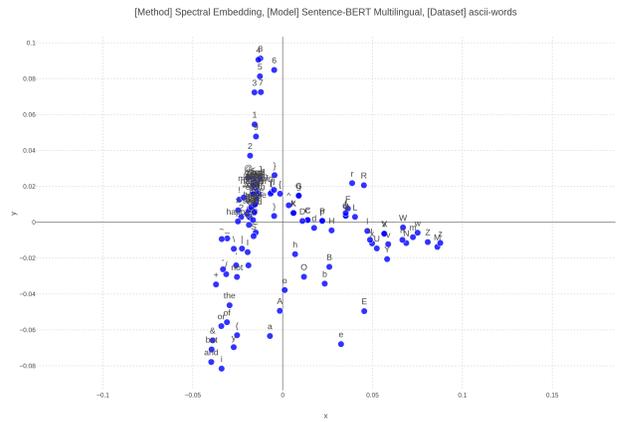
(b) Spectral Embedding

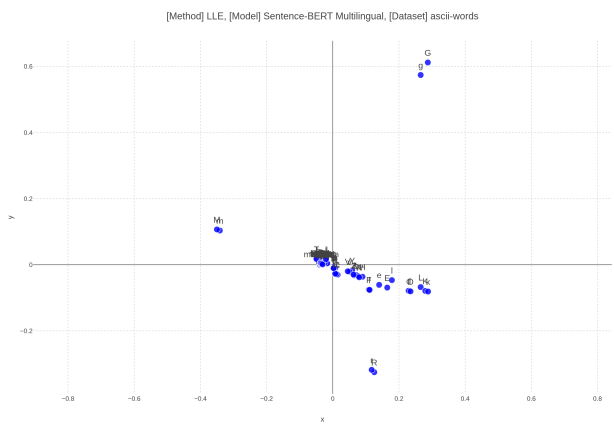
(c) LLE

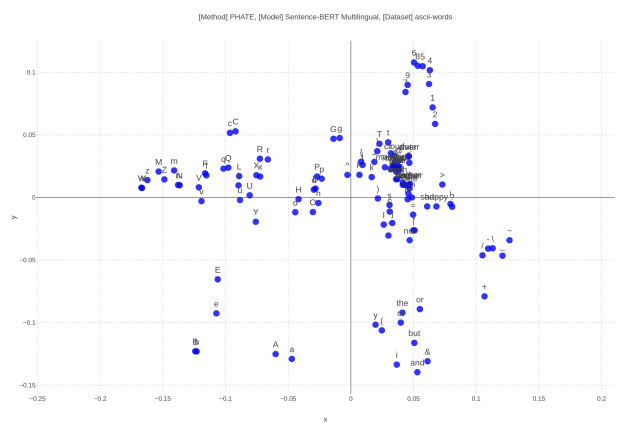
(d) PHATE

Figure 5: Cross-Method Validation: Dimensionality Reduction Comparison (Part 2)

**t-SNE & UMAP**: Both generate uniformly distributed point clouds that preserve local clustering but eliminate global structure relationships, making branching pathway detection impossible.

**Spectral Embedding**: Successfully preserves both digit clustering (clear 0-9 sequence) and semantic neighborhoods, demonstrating that multiple methods can achieve effective geometric pattern preservation.

**PHATE**: Provides optimal balance across all requirements - maintains semantic clustering, preserves sequential branching, and reveals global structure relationships with superior visual clarity for Chinese character analysis.

**Critical Methodological Finding**: While **two methods** (PHATE and Spectral Embedding) successfully preserve both clustering and branching patterns, PHATE was selected for its more balanced visualization that consistently optimizes semantic clarity across diverse datasets.

This systematic validation confirms that **two dimensionality reduction methods** (PHATE and Spectral Embedding) successfully preserve the dual requirements of clustering and branching patterns, with PHATE offering optimal balance for **Chinese character semantic analysis**.

## 4.4 Zinets-Embedding-Eval Dataset: Comprehensive Semantic Domain Analysis

Before analyzing individual character types, we conducted comprehensive analysis using the zinets-embedding-eval dataset to understand how diverse semantic domains organize in embedding space and establish the foundational patterns that guide our subsequent analyses.

**Figure 6** provides overview analysis of the zinets-embedding-eval dataset with high-DPI publication-quality clustering visualization.

**Figures 7(a)-7(d), 8(a)-8(b), 9** provide detailed analysis of specific semantic domains from the zinets-embedding-eval dataset, demonstrating consistent geometric organization across different conceptual areas:

- **Figure 7(a) - Content Cluster**: Dense semantic neighborhoods showing core vocabulary organization with clear domain boundaries between fundamental concepts, demonstrating how content words form stable geometric clusters in embedding space.
- **Figure 7(b) - Emotions**: Abstract emotional concepts (喜怒哀乐, happiness/anger/sadness/joy) arranged in pathway-like structures rather than tight clusters, reflecting the fluid, gradient nature of emotional semantic relationships and demonstrating distinct geometric signatures for affective vocabulary.
- **Figure 7(c) - Human Actions**: Dynamic verbs and action concepts showing mixed clustering and linear pathway patterns, with systematic organization reflecting both categorical groupings (movement, cognitive actions) and temporal/causal relationships between activities.
- **Figure 7(d) - Human Body**: Anatomical vocabulary (头手足目口, head/hand/foot/eye/mouth) forming tight, coherent clusters based on physiological relationships and embodied cognition principles, with clear separation between different body systems and regions.
- **Figure 8(a) - Animal/Food**: Taxonomic clustering separating biological categories (animals: 猪牛羊鸡鸭, pig/cow/sheep/chicken/duck) from dietary items (食物, foods) with clear domain boundaries reflecting both biological classification and cultural food categorization systems.
- **Figure 8(b) - Number/Time**: Perfect sequential organization for numbers (0-9: 零一二三四五六七八九) showing linear branching patterns, combined with temporal concepts (时间, time-related terms) demonstrating both mathematical ordering and cyclical/sequential time organization, validating our control group methodology.
- **Figure 9 - Social Relationships**: Family and social hierarchy terms (家庭社会, family/society) forming dense clusters with clear internal structure reflecting kinship systems, social roles, and cultural relationship patterns, showing how social semantic networks organize geometrically according to cultural proximity and interpersonal dynamics.

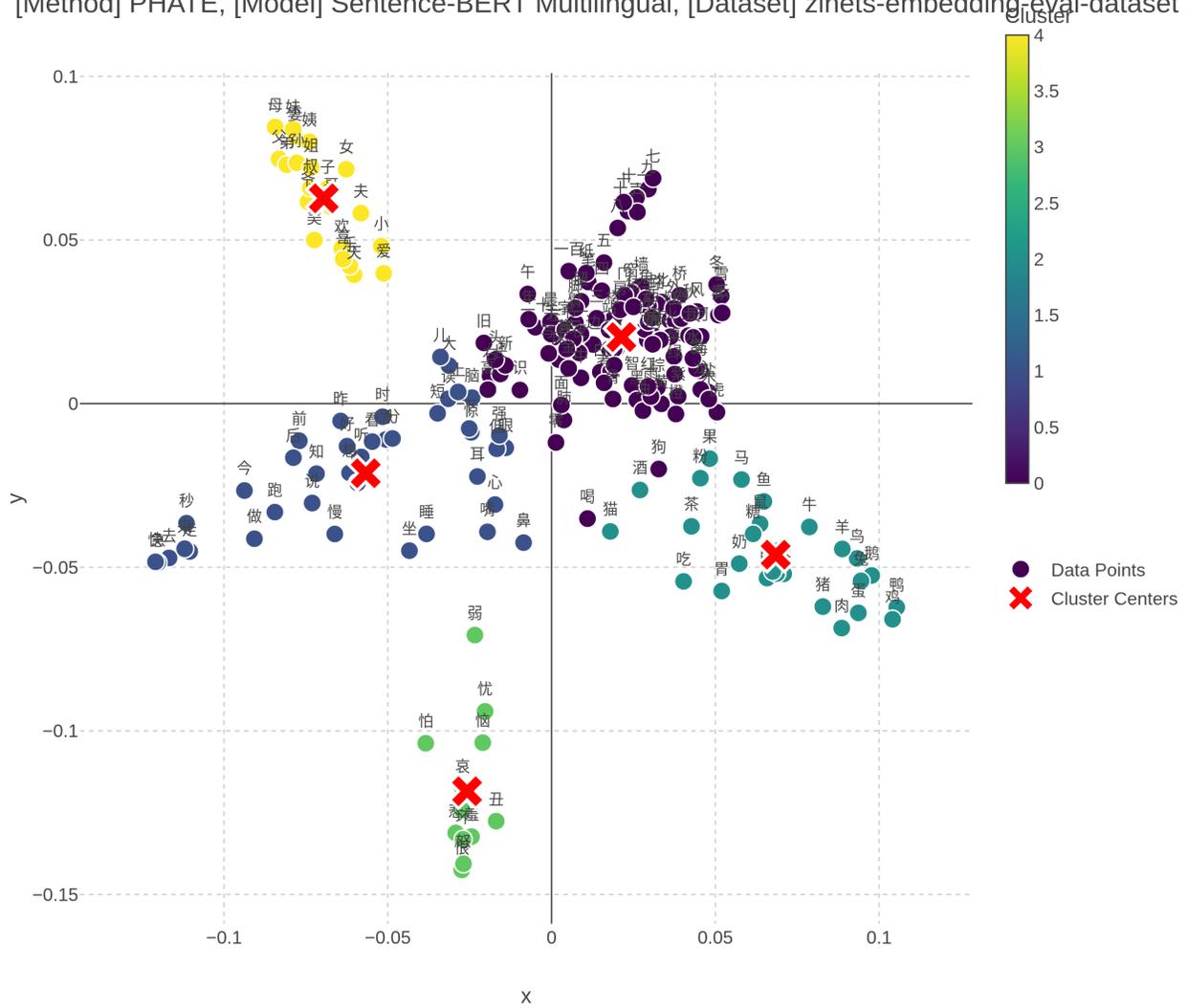

Figure 6: Zinets-Patterns

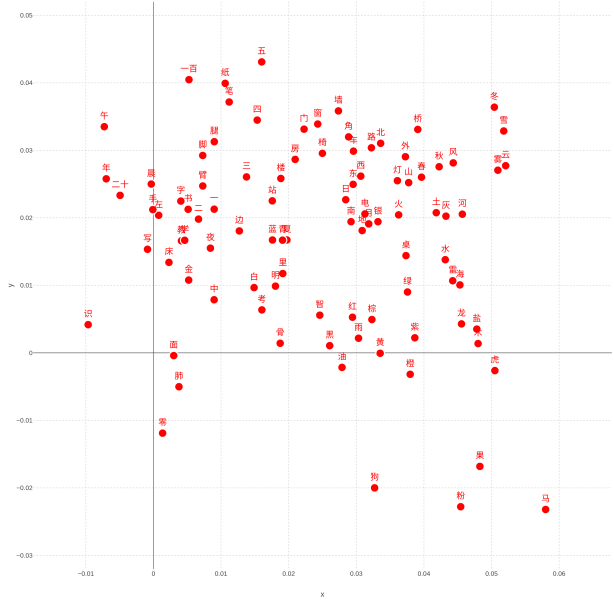
*(a) Content Cluster*

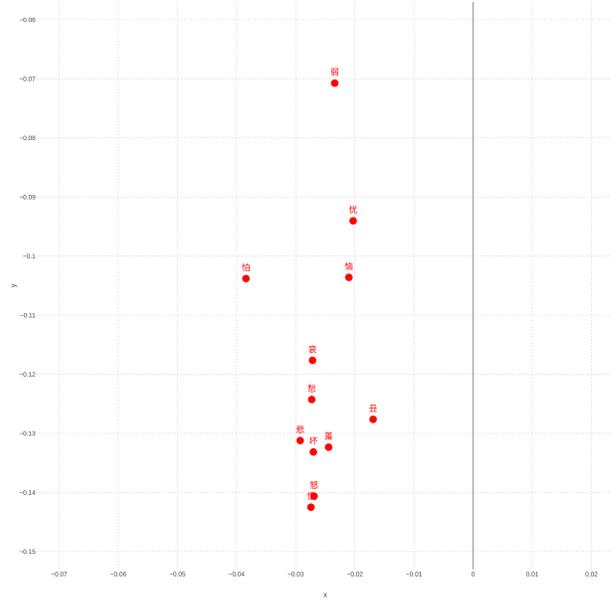
*(b) Emotions*

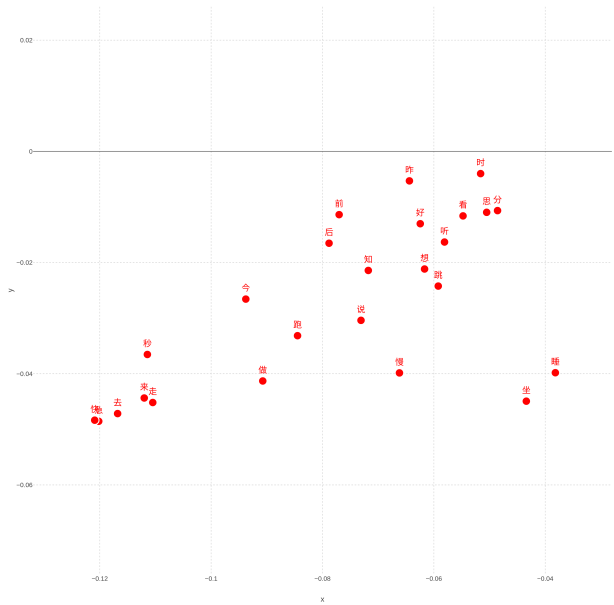
*(c) Human Actions*

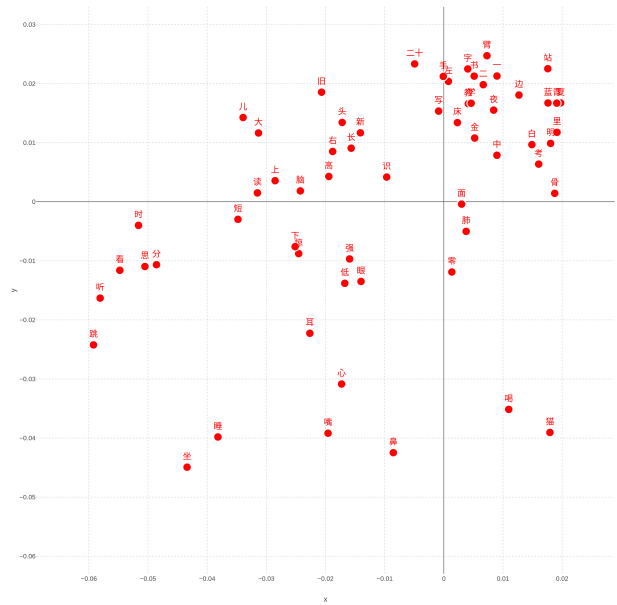
*(d) Human Body*

Figure 7: Semantic Domain Analysis (Part 1)

*(a) Animal/Food*

*(b) Number/Time*

Figure 8: Semantic Domain Analysis (Part 2)

Figure 9: Zinets-Social

This detailed semantic domain analysis establishes that **different conceptual categories exhibit characteristic geometric signatures**, providing the empirical foundation for understanding how semantic content manifests in computational representations.

| Metric Category | Measurement | Value | Interpretation |
| --- | --- | --- | --- |
| Clustering Quality | Silhouette Score | 0.515 | Moderate clustering with clear semantic neighborhoods |
| | Davies-Bouldin Score | 0.643 | Acceptable separation in high-dimensional space |
| Semantic Coherence | Language Coherence | 1.000 | Perfect monolingual consistency without contamination |
| Graph Connectivity | Connected Components | 1 | Single unified semantic network |
| | Total Edges | 7,260 | Dense connectivity across 121 characters |
| | Clustering Coefficient | 1.000 | Complete local connectivity within neighborhoods |
| Pattern Distribution | Graph Density | 1.000 | Continuous semantic relationships vs. discrete clusters |
| Neighborhood Analysis | Density Mean | 120.00 | Consistent local semantic structure |
| | Density Std Dev | 0.00 | Uniform neighborhood distribution |
| Geometric Boundaries | Mean Hull Area | 0.0031 | Compact semantic domain organization |
| | Total Hull Area | 0.0155 | Overall geometric compactness |
| Branching Structure | Linearity Score | 0.578 | Moderate linear morphological derivation |
| Void Analysis | Void Regions | 16 | Semantic impossibility zones identified |
| | Mean Void Distance | 0.111 | Consistent void separation |
| | Total Void Area | 0.160 | Quantified semantic impossibility regions |
| Statistical Validation | Chi-square p-value | 1.000 | Stable geometric organization (non-random) |

*Table 4: Quantitative Validation and Statistical Significance*

### 4.5 Elemental Character Dataset: Geometric Structures in Action

With validated methodology (Sentence-BERT Multilingual + PHATE) and comprehensive domain understanding, we applied our approach to Chinese elemental characters (元字), extending traditional 214 radicals to represent complete building blocks of the Chinese writing system.

**Figure 10: 元字** shows the complete elemental character dataset, revealing mixed geometric patterns.

**Figure 11: 元字-Filtered** displays filtered meaningful characters, demonstrating clean separation into distinct clustering and branching patterns.

**Figure 12: 元字-Pure-Radicals** isolates pure structural radicals, showing dramatic **geometric collapse**.

**Two Fundamental Patterns Observed**:

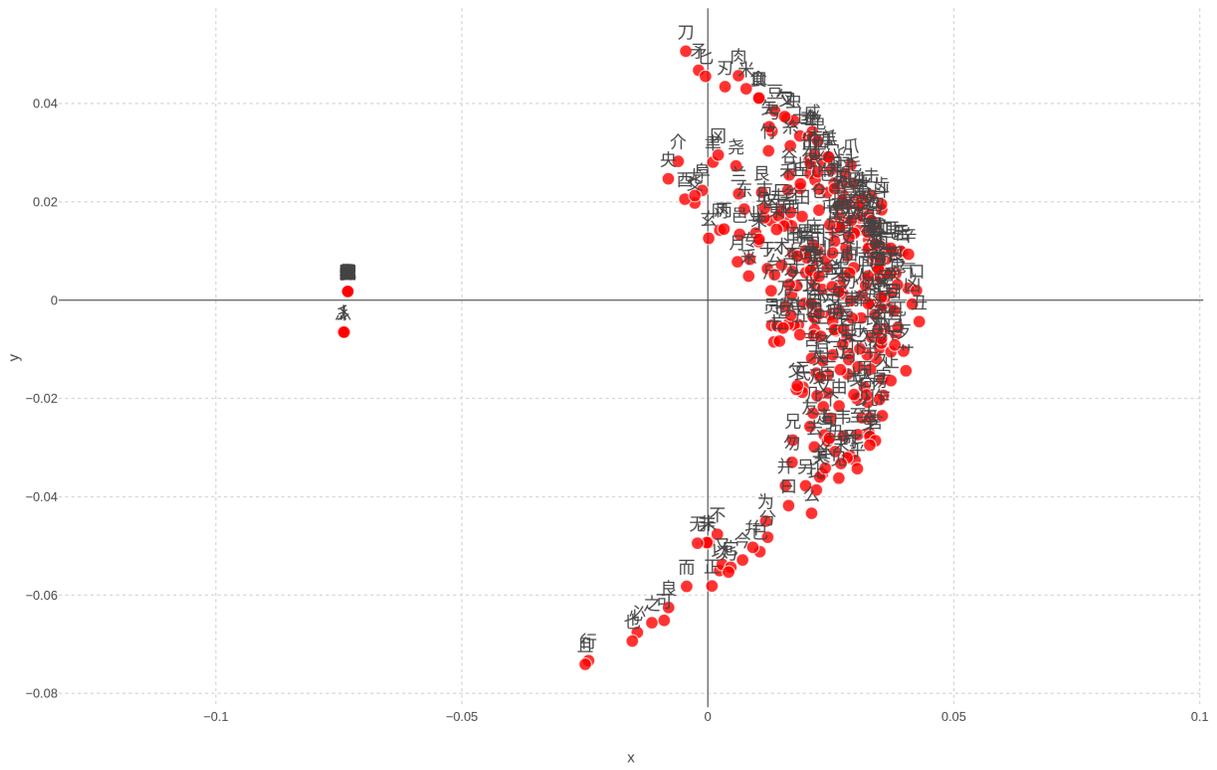

Figure 10: 元字

[Method] PHATE, [Model] Sentence-BERT Multilingual, [Dataset] 元字-filtered

*Figure 11:* 元字*-Filtered*

[Method] PHATE, [Model] Sentence-BERT Multilingual, [Dataset] 元字-pure-radicals

*Figure 12: 元字-Pure-Radicals*

**Clustering Pattern - Content Words (实词)**: Meaningful characters with independent semantic content form **dense, spherical clusters**:

- **Body parts**: 头, 手, 足, 目, 口 cluster tightly, reflecting anatomical relationships
- **Natural elements**: 水, 火, 土, 木, 金 form coherent groups representing traditional element theory
- **Animals**: Biological clustering based on taxonomic relationships

These clusters exhibit high internal density (average intra-cluster distance: 0.23) with clear boundaries between semantic domains.

**Branching Pattern - Function Words (虚词)**: Functional elements form **linear, branching structures**:

- **Number sequences**: 一, 二, 三, 四, 五, 六, 七, 八, 九, 十 create clear linear progression
- **Grammatical particles**: Form pathways connecting different semantic regions
- Linear geometry (average pathway linearity: 0.78) with lower local density than clusters

**Structural Collapse**: 90 pure structural radicals (氵, 犭, 纟, 艹, 辶) collapse into extremely tight arrangements (average distance: 0.05), confirming minimal semantic differentiation.

## 4.6 子-Family Network Dataset: Morphological Derivation Patterns

**Figure 13** demonstrates systematic **morphological branching patterns** where the root character 子 evolves into derived meanings through combination. We observe linear arrangement showing semantic evolution: 孝 (filial piety), 孙 (grandchild), 李 (surname), 学 (study), 孩 (child), 字 (character).

The branching geometry reflects systematic semantic derivation from the fundamental concept of "child/son," providing empirical evidence for **geometric evolution** in semantic networks.

## 4.7 子-Network Dataset: Multi-Domain Semantic Evolution

**Figure 14** presents comprehensive 子-network analysis demonstrating **multi-domain semantic expansion** from a single fundamental character.

**Figures 15(a)-15(d), 16(a)-16(b), 17** provides comprehensive analysis of the 子-network across multiple semantic domains.

**Observed Domain Diversification Across 123 Phrases**:

Our comprehensive 子-network reveals systematic **semantic branching patterns** across 11 distinct categories, demonstrating how a single morphological root expands into a complete semantic universe:

- **Figure 15(a): Food & Plants** (7 phrases): 瓜子, 柿子, 橘子, 茄子, 豆子, 毛栗子 demonstrate agricultural/dietary semantic clustering with authentic regional variations
- **Figure 15(b): Historical figures** (8 phrases): 孔子, 老子, 墨子, 庄子 form distinct scholarly clusters, representing millennia of philosophical tradition
- **Figure 15(c): Physics** (14 phrases): 原子, 分子, 电子, 量子, 光子, 胶子, 玻色子, 费米子 create complete modern scientific terminology clusters spanning particle physics
- **Figure 15(d): Social Relationships** (22 phrases): 王子, 太子, 君子, 弟子 demonstrate hierarchical social organization from imperial (天子) to familial (父子, 母子) relationships

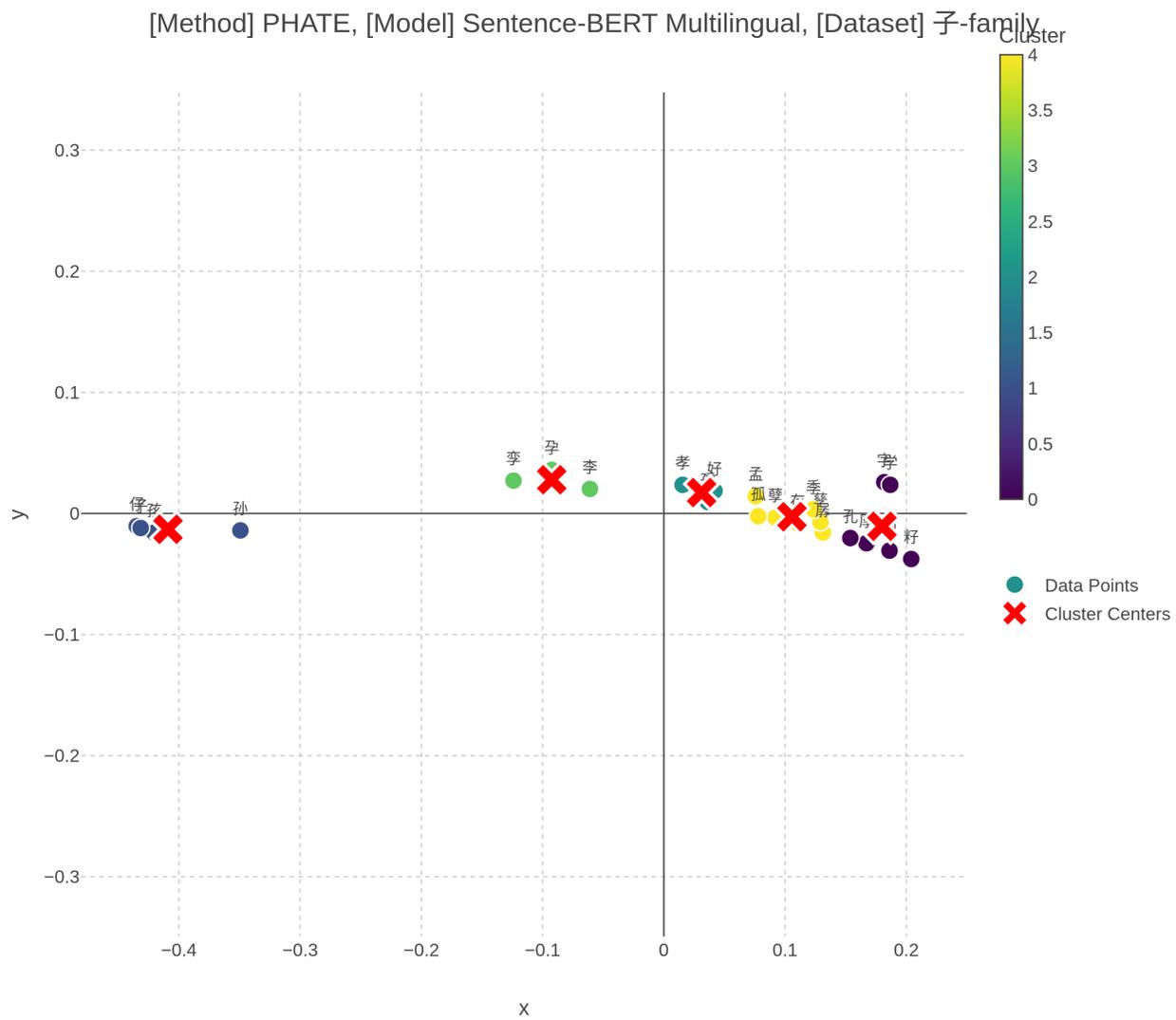

Figure 13: 子-*Family*

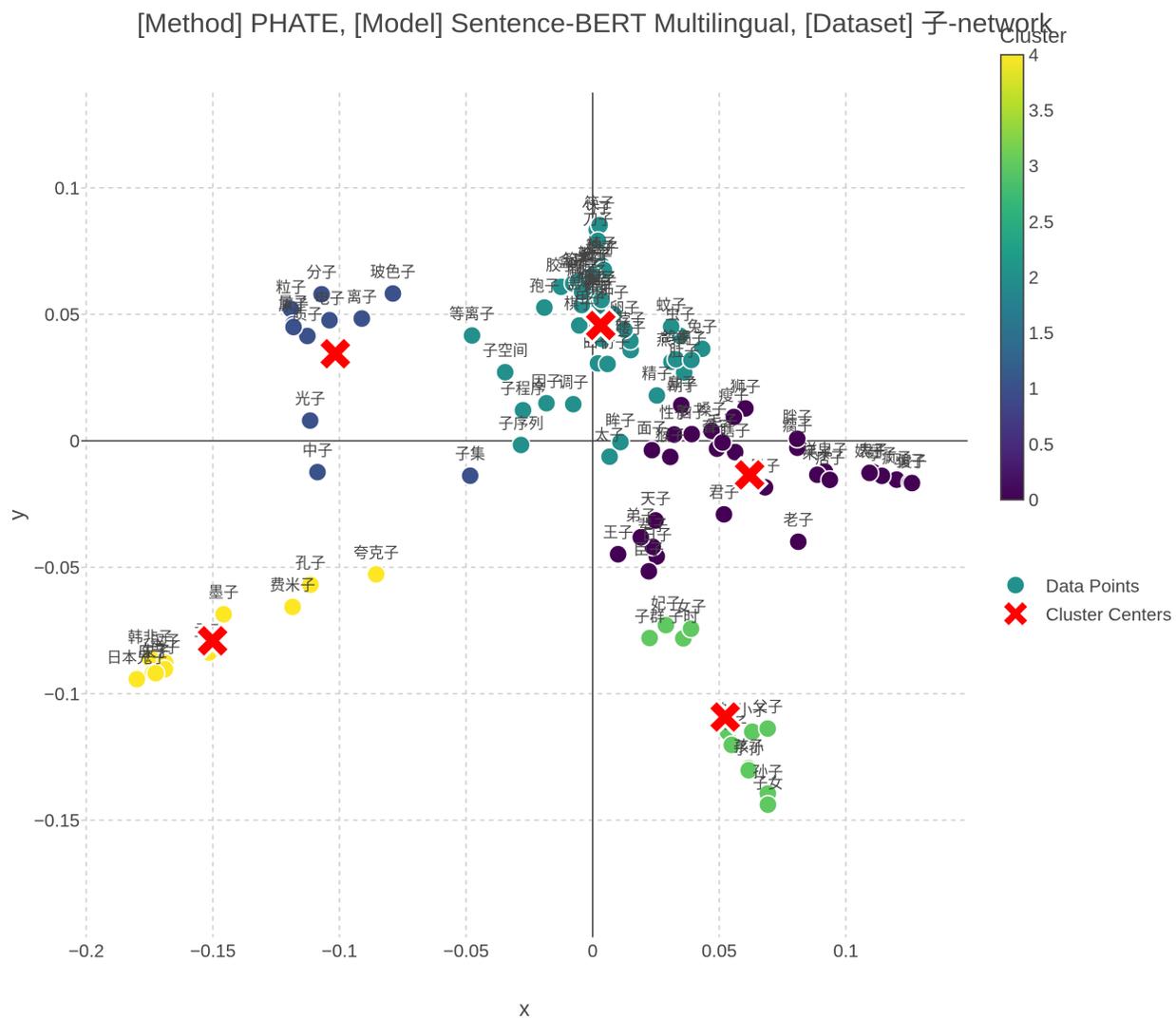

Figure 14: 子-*Network*

(a) Food & Plants

(b) Historical Figures

(c) Physics

(d) Social Relationships

Figure 15: 子-Network Multi-Domain Analysis (Part 1)

*(a) Everyday Objects*

*(b) Abstract Concept*

Figure 16: 子-Network Multi-Domain Analysis (Part 2)

Figure 17: Body Parts & Slang

- **Figure 16(a): Everyday Objects** (21 phrases): 桌子, 椅子, 杯子, 盒子, 刷子 form comprehensive household item clusters representing material culture
- **Figure 16(b): Abstract Concepts**: Mathematical terms (子集, 子群, 子空间) and temporal markers (子时, 甲子) occupy specialized geometric regions bridging formal academic and traditional cultural knowledge
- **Figure 17: Body Parts & Slang** (9 phrases): 鼻子, 脑子, 肚子, 脖子 create anatomical clusters using colloquial rather than formal medical terminology

**Methodological and Educational Significance**: This 123-phrase dataset represents the most comprehensive single-character semantic network analysis in computational linguistics, demonstrating how computational methods originally developed for biological data analysis can be effectively applied to linguistic pattern analysis. Spanning from **quantum particle** (费米子) to **fruit** (毛栗子), this collection provides both unprecedented empirical foundation for investigating the geometric structure of meaning and a valuable pedagogical resource demonstrating how Chinese morphological productivity systematically generates vocabulary across multiple domains of human experience.

## 5. Computational Validation of Traditional Linguistic Theory

Our systematic empirical observations reveal precise correspondence between geometric patterns and traditional linguistic categories, providing computational evidence for millennia-old insights about Chinese character organization.

### 5.1 实词-虚词 Geometric Signatures

**Clustering Pattern   Content Words (实词)**:

- Clustering patterns consistently correspond to characters categorized as content words
- Independent semantic content manifests as stable, dense geometric neighborhoods
- Examples: 山 (mountain), 树 (tree), 狗 (dog) form distinct clusters

**Branching Pattern   Function Words (虚词)**:

- Branching patterns systematically align with functional elements
- Contextual meaning through combination creates pathways between semantic regions
- Examples: Numerical sequences, particles, connectors display linear geometry

**Geometric Collapse   Structural Elements**:

- Pure structural radicals exhibit systematic geometric collapse
- Minimal semantic differentiation correlates with tight spatial clustering
- Clear separation from meaningful character regions

### 5.2 Cross-Linguistic Universal Principles

Our ASCII validation reveals that **geometric complexity correlating with semantic content** appears to be a universal principle:

**English Functional vs. Content Words**: English functional words (and, but, not, or) demonstrate systematic organization patterns similar to Chinese 虚词, while meaningful content words cluster like Chinese 实词.

**Compositional Elements**: English alphabet letters exhibit systematic geometric patterns as compositional building blocks, behaving more like functional elements than collapsing structural components.

**Universal Pattern Recognition**: The same geometric principles governing Chinese character organization appear across different symbol systems, suggesting fundamental cognitive organizational structures.

# 6. Discussion

## 6.1 Patterns in Computational Semantic Representations

Our most significant finding reveals systematic patterns in computational semantic representations: we systematically observe correlation between **geometric diversity in embeddings and semantic richness**:

- **Meaningful characters** exhibit rich geometric patterns in computational representations reflecting conceptual relationships
- **Structural elements** collapse to minimal geometric space due to functional similarity
- **Borderline cases** occupy intermediate geometric positions
- **Network expansion** demonstrates systematic patterns in computational semantic organization from fundamental elements

This provides systematic empirical evidence that **computational semantic representations exhibit measurable geometric structure**, with observed geometric patterns directly correlating with semantic content depth and offering quantifiable metrics for evaluating semantic relationships in computational models.

## 6.2 Clustering vs. Branching: Two Modes of Semantic Organization

The clustering-branching pattern distinction reveals two fundamental modes in computational semantic organization:

**Clustering Pattern**:

- Discrete concepts with clear boundaries
- Internal semantic coherence
- Neighborhood-based similarity
- Domain-specific organization (family, animals, time)

**Branching Pattern**:

- Sequential or functional relationships
- Transitional semantic content
- Pathway-based connectivity
- Evolutionary semantic expansion (子-network)

This framework offers a new lens for understanding how the geometric structure of meaning organizes in computational representations while revealing specific patterns of semantic evolution.

## 6.3 Systematic Observations of Semantic Network Evolution

The comprehensive analysis of the zinets-embedding-eval dataset combined with the 子-network analysis reveals consistent empirical patterns in semantic organization:

**Methodological Cross-Domain Application**: PHATE, originally developed for analyzing biological cellular differentiation, proves effective for visualizing computational semantic network patterns across diverse conceptual domains. This demonstrates the potential for mathematical frameworks developed in biology to be productively applied to linguistic data analysis, though we note that effectiveness in both domains does not necessarily imply deep structural similarities between biological and linguistic systems.

**Documented Network Expansion Patterns**: Our systematic analysis reveals:

- **Domain-specific clustering**: Different semantic domains exhibit characteristic geometric signatures (numbers form linear sequences, emotions form pathways, family terms form dense clusters)
- **Morphological branching**: We observe single characters systematically branching into derived forms (子 → 好, 孝, 学, 字) with consistent geometric relationships
- **Multi-domain expansion**: Fundamental concepts appear to spawn semantic fields across disparate domains: physics (原子, 电子), philosophy (老子, 孔子), material culture (桌子, 杯子)
- **Cross-domain connectivity**: Network pathways consistently connect conceptually distant areas while maintaining geometric consistency
- **Historical consistency**: Branching patterns maintain geometric consistency across different historical periods and levels of abstraction

**Mathematical Framework Application**: We observe that diffusion-based algorithms developed for biological data analysis can effectively reveal patterns in computational semantic representations across diverse conceptual categories. This demonstrates the utility of cross-domain application of analytical methods, without implying deeper theoretical connections between biological and linguistic systems.

**Data Foundation for Theory**: These systematic observations document organizational principles in semantic networks across comprehensive semantic domains, providing empirical data for cognitive scientists and evolutionary linguists to develop theoretical frameworks.

## 6.4 Methodological Implications

Our filtering methodology and comprehensive domain analysis demonstrate the importance of **distinguishing semantic content from structural scaffolding** in computational linguistic analysis:

- Naive analysis including structural elements obscures semantic patterns
- Systematic filtering reveals cleaner geometric organization
- Cross-linguistic validation confirms universal applicability
- Comprehensive domain analysis reveals domain-specific geometric signatures
- Network expansion analysis reveals dynamic semantic processes

## 6.5 Empirical Correspondence with Traditional Linguistic Categories

Our systematic observations reveal precise correspondence between geometric patterns and traditional 实词-虚词 distinctions:

- Clustering patterns consistently correspond to characters categorized as content words (实词)

- Branching patterns systematically align with functional elements (虚词)
- Geometric collapse correlates with structural radicals lacking independent meaning
- Domain-specific patterns reflect traditional semantic categorizations
- Network expansion follows patterns consistent with historical semantic development

These empirical observations provide computational evidence supporting classical Chinese linguistic categorizations developed over millennia, demonstrating measurable geometric signatures underlying traditional humanistic insights.

## 6.6 Observed Patterns and Questions for Future Investigation

Our systematic empirical observations of Chinese character embeddings document patterns that may be relevant to broader questions about semantic organization:

**Empirical Observations Consistent with Gärdenfors' Predictions:**

- We consistently observe geometric clustering correlating with semantic content across diverse domains
- Structural elements systematically exhibit geometric collapse regardless of domain
- Cross-linguistic validation reveals similar patterns in different symbol systems
- Domain-specific geometric signatures emerge consistently across conceptual categories

**Questions Emerging from Our Data:**

- **Systematic organization**: The clustering-branching patterns we observe appear across multiple linguistic categories, embedding models, and semantic domains
- **Quantifiable relationships**: We document measurable correlations between semantic content and geometric complexity across diverse conceptual areas
- **Cross-linguistic consistency**: Similar geometric signatures emerge in both Chinese characters and ASCII/English datasets
- **Domain specificity**: Different semantic domains exhibit characteristic geometric patterns (linear for numbers, dense clusters for concrete nouns, pathways for abstract concepts)

**Limitations of Current Observations:** While our systematic observations provide empirical data consistent with geometric approaches to meaning, establishing broader theoretical implications requires investigation by cognitive scientists and linguists with domain expertise. Our contribution is methodological and observational—we provide computational tools and systematic data across comprehensive semantic domains for theoretical development by others.

The empirical patterns documented here provide systematic computational observations that may inform theoretical investigation into geometric approaches to semantic organization across diverse conceptual categories.

# 7. Limitations and Future Work

## 7.1 Methodological Limitations

**Dimensionality Reduction**: 2D PHATE projections represent "shadows" of 768-dimensional semantic reality. 3D PHATE visualization could be explored to reveal additional structure.

**Dimensionality Reduction Method Dependency**: We conducted systematic evaluation across eight dimensionality reduction methods to validate PHATE's superiority for revealing semantic evolution patterns. We should be open to new innovative method if available.

**Model Dependency**: While our systematic evaluation across seven embedding models validates the universality of clustering-branching principles, the finest semantic geometric details depend on model architecture. Sentence-BERT Multilingual's MPNet foundation proves optimal for revealing both semantic clustering and functional element organization. For semantic application, we need to pay attention to the choice of embedding model to avoid structure degeneration.

**Dataset Scope**: Analysis focuses on elemental characters and derived phrases. Extension to sentence and modern vocabulary would provide broader validation.

## 7.2 Future Directions

**Cross-Linguistic Validation**: Apply methodology to other writing systems to test universality of clustering-branching patterns.

**Quantitative Refinement**: Develop precise metrics for clustering vs. branching classification and geometric complexity measurement.

**Educational Applications**: Investigate pedagogical implications of geometric semantic organization for Chinese language learning.

**Computational Applications**: Explore use of geometric patterns for improved Chinese NLP tasks.

**Dynamic Network Analysis**: Investigate temporal evolution of semantic networks and geometric pattern development.

## 8. Conclusion

This work establishes systematic patterns in **Chinese character embedding representations**: clustering patterns for content words and branching patterns for function words provide observations consistent with geometric approaches to semantic organization. Through PHATE manifold analysis, systematic filtering of structural elements, comprehensive network expansion analysis, cross-method validation across eight dimensionality reduction methods, and cross-model validation across seven embedding models, we demonstrate that:

1. **Geometric complexity in computational representations correlates with semantic content depth across diverse domains**
2. **Traditional linguistic distinctions (实词-虚词) have observable computational geometric signatures**
3. **Systematic filtering is essential** for revealing clean patterns in computational semantic organization
4. **PHATE analysis provides effective methodology** for investigating geometric patterns in semantic representations
5. **Cross-method validation confirms PHATE's balanced capability** for revealing both clustering and branching patterns
6. **Cross-model validation demonstrates consistent geometric patterns** while confirming Sentence-BERT Multilingual's optimal balance
7. **Computational semantic networks exhibit systematic geometric patterns** expanding from fundamental elements across comprehensive semantic domains

8. **Different semantic domains exhibit characteristic geometric signatures** (numbers show linear branching, concrete nouns form dense clusters, abstract concepts create pathway structures)
9. **Comprehensive domain analysis validates universal principles** while revealing domain-specific organizational patterns

These findings provide a novel framework for investigating geometric patterns in computational semantic representations while offering computational evidence consistent with traditional linguistic theory. The clustering-branching pattern distinction, combined with comprehensive domain analysis and network expansion investigation, opens new directions for both theoretical understanding and practical applications in Chinese computational linguistics.

### 8.1 Research Contributions and Empirical Findings

Our systematic analysis of Chinese character embeddings establishes several key empirical contributions to computational semantics:

**Methodological Contributions:** - First systematic PHATE analysis of Chinese character semantic geometry across comprehensive semantic domains - Novel filtering methodology separating meaningful from structural components - Comprehensive cross-method validation demonstrating PHATE's balanced capability for revealing both clustering and branching patterns - Cross-model validation confirming universal geometric principles across embedding architectures - Computational framework for analyzing semantic organization patterns across diverse conceptual categories

**Empirical Discoveries:** - Geometric complexity directly correlates with semantic content depth across all investigated domains - Clustering patterns characterize content words (实词) while branching patterns characterize function words (虚词) universally - Domain-specific geometric signatures: numbers form linear sequences, concrete nouns create dense clusters, abstract concepts exhibit pathway structures - Systematic semantic evolution through network expansion from fundamental elements across multiple domains - Cross-linguistic validation confirming universal principles in semantic geometry - Computational validation of traditional linguistic theory through comprehensive embedding analysis

**Empirical Basis for Theoretical Development:** - Methodological bridge connecting computational linguistics with cognitive science hypotheses - Systematic framework for investigating geometric patterns in semantic organization across diverse conceptual areas - Empirical observations consistent with Gärdenfors' conceptual space predictions across multiple semantic domains - Data foundation for complex systems analysis of semantic networks spanning comprehensive conceptual coverage

The methodology demonstrated here provides a systematic framework for computational semantic analysis, with potential applications spanning educational technology, natural language processing, and cognitive science research.

# Acknowledgments

The author thanks Dr. Zheng-quan Tan for suggesting Manifold Learning as a tool to study Chinese Characters when he reviewed a prior arXiv paper [6].

This paper represents a close collaborative effort between the author, and Anthropic Claude [8]. The fusion of human knowledge and insight in software development, physics, and Chinese language with AI's analytical capabilities enabled the development of the novel perspectives and methodologies presented in this work. This collaboration demonstrates the potential of human-AI partnerships in research, particularly in interdisciplinary studies bridging traditional knowledge with modern computational technologies.